\documentclass[letterpaper]{article} 
\usepackage{aaai2026}  
\usepackage{times}  
\usepackage{helvet}  
\usepackage{courier}  
\usepackage[hyphens]{url}  
\usepackage{graphicx} 
\usepackage{natbib}  
\usepackage{caption} 
\usepackage{algorithm}
\usepackage{algorithmic}
\usepackage{threeparttable}
\usepackage{multirow}
\usepackage{booktabs}
\usepackage{array}

\usepackage{newfloat}
\usepackage{listings}
\usepackage{amsmath}
\usepackage{amssymb}
\usepackage{bm}
\DeclareCaptionStyle{ruled}{labelfont=normalfont,labelsep=colon,strut=off} 
\floatstyle{ruled}
\newfloat{listing}{tb}{lst}{}
\floatname{listing}{Listing}
\title{Divide-and-Conquer Decoupled Network for\\Cross-Domain Few-Shot Segmentation}
\author{
    Runmin Cong\textsuperscript{\rm 1,\rm 2}, 
    Anpeng Wang\textsuperscript{\rm 1},
    Bin Wan\textsuperscript{\rm 1}\thanks{Corresponding author.},
    Cong Zhang\textsuperscript{\rm 3},
    Xiaofei Zhou\textsuperscript{\rm 4},
    Wei Zhang\textsuperscript{\rm 1}
}
\affiliations{
    \textsuperscript{\rm 1}School of Control Science and Engineering, Shandong University, China\\
    \textsuperscript{\rm 2}State Key Laboratory of Autonomous Intelligent Unmanned Systems, China\\
    \textsuperscript{\rm 3}Department of Electrical and Electronic Engineering, The Hong Kong Polytechnic University, China\\
    \textsuperscript{\rm 4}School of Automation, Hangzhou Dianzi University, China\\

    rmcong@sdu.edu.cn, rawwap@mail.sdu.edu.cn, binwan@sdu.edu.cn,\\
    cong-clarence.zhang@connect.polyu.hk, zxforchid@hdu.edu.cn, davidzhang@sdu.edu.cn
}

\usepackage{bibentry}

\begin{document}

\maketitle

\begin{abstract}
Cross-domain few-shot segmentation (CD-FSS) aims to tackle the dual challenge of recognizing novel classes and adapting to unseen domains with limited annotations. However, encoder features often entangle domain-relevant and category-relevant information, limiting both generalization and rapid adaptation to new domains. To address this issue, we propose a Divide-and-Conquer Decoupled Network (DCDNet). 
In the training stage, to tackle feature entanglement that impedes cross-domain generalization and rapid adaptation, we propose the Adversarial-Contrastive Feature Decomposition (ACFD) module. It decouples backbone features into category-relevant private and domain-relevant shared representations via contrastive learning and adversarial learning. 
Then, to mitigate the potential degradation caused by the disentanglement, the Matrix-Guided Dynamic Fusion (MGDF) module adaptively integrates base, shared, and private features under spatial guidance, maintaining structural coherence. 
In addition, in the fine-tuning stage, to enhanced model generalization, the Cross-Adaptive Modulation (CAM) module is placed before the MGDF, where shared features guide private features via modulation ensuring effective integration of domain-relevant information.
Extensive experiments on four challenging datasets show that DCDNet outperforms existing CD-FSS methods, setting a new state-of-the-art for cross-domain generalization and few-shot adaptation. Code: https://github.com/rawwap/DCDNet.
\end{abstract}


\begin{figure}[!t]
\centering
\includegraphics[width=1.0\linewidth]{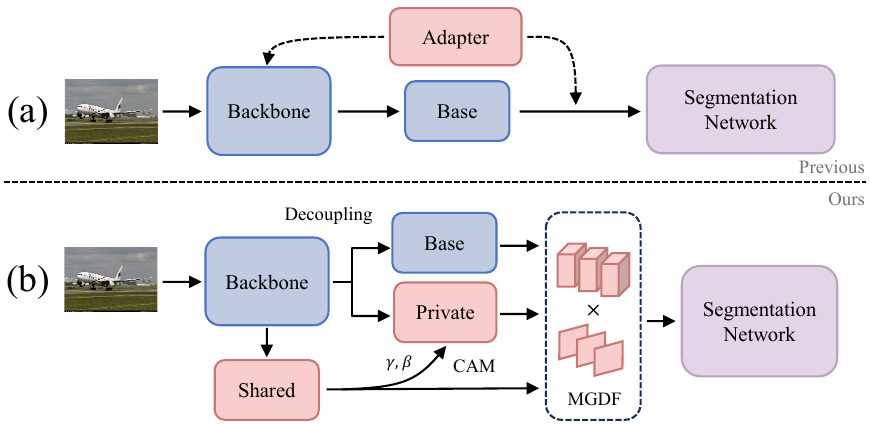}
\caption{\textbf{Comparison of existing CD-FSS methods and ours.} (a) Most existing methods introduce adapter-like modules to backbone architectures or feature embeddings, yet neglect to address the intrinsic information redundancy within base features themselves. (b) Our method decomposes and refines the base features, then effectively integrates these distilled features to endow the model with enhanced generalization and domain adaptation capabilities.}
\label{motivation}
\end{figure}

\section{Introduction}

Semantic segmentation, as a core task in visual scene understanding, aims to assign category labels to each pixel in an image and supports downstream applications such as autonomous driving, medical analysis, and visual navigation~\cite{long2015fully, zhou2017scene}. While recent methods have made substantial progress, their success critically depends on large-scale, densely annotated datasets~\cite{sun2022coarse}, which are expensive and labor-intensive to acquire – particularly for rare or emerging categories in open-world scenarios. To address this limitation and enhance model adaptability in dynamic, open-world scenarios where novel categories frequently emerge, Few-Shot Semantic Segmentation (FSS) has gained growing interest~\cite{vinyals2016matching, liu2022dynamic, moon2023msi, cong2024query, xiong2025mm, luo2025concept}. Despite their promise, current FSS models are typically developed with both training and testing conducted within the same domain~\cite{peng2024multi}. However, in real-world applications, the test domain often differs significantly from the training domain~\cite{lei2022cross}. For example, a model trained on natural scene images is often expected to perform well on entirely different domains such as medical or remote sensing imagery. This domain gap poses a major challenge to the robustness and generalization ability of existing FSS methods, giving rise to the more demanding task of Cross-Domain Few-Shot Semantic Segmentation (CD-FSS), where models must handle both novel categories and domain shifts using only a few annotated examples \cite{lei2022cross}.
For CD-FSS tasks, existing approaches often introduce adapter-like modules to subtly adjust the backbone or intermediate features \cite{su2024domain, tong2024lightweight, liu2025devil}, as shown in Figure \ref{motivation}(a), aiming to extract domain-invariant representations and enhance generalization to unseen domains. However, the features extracted by the backbone are typically entangled with multiple types of object information (\emph{i.e.,} domain information and category information). Directly feeding these entangled features into the adapter and segmentation modules not only limits the model’s generalization capability but also hinders effective transfer and rapid adaptation to new domains.

To address the aforementioned challenge, we propose the Divide-and-Conquer Decoupled Network (DCDNet), a framework that decomposes feature representations into domain-relevant shared features that capture low-level structural information and category-relevant private features that encode high-level semantics, as illustrated in Figure \ref{motivation}(b). On the one hand, in the training phase, encoder features are entangled with domain and category information, leading to poor adaptability to new domains, so we design the Adversarial-Contrastive Feature Decomposition (ACFD) module. It applies multiple convolutions to high-level and low-level features respectively to obtain category-relevant private features and domain-relevant shared features. However, convolutions cannot precisely decouple intrinsically entangled base features, prompting us to design specialized learning mechanisms: for high-level private features, ACFD uses contrastive learning ~\cite{khosla2020supervised} to enhance intra-class compactness and inter-class separability (boosting private feature category purity), while for low-level shared features, ACFD uses adversarial learning ~\cite{ebrahimi2020adversarial} to help the model to capture category-agnostic commonalities. After that, due to mitigate the degradation of structural integrity and stability when disentangling encoder features, we design a Matrix-Guided Dynamic Fusion (MGDF) module which integrates the base features, domain-relevant shared features, and category-relevant private features, while introducing a spatially-aware matrix to guide their dynamic fusion in a more adaptive and coherent manner. 
On the other hand, in the fine-tuning phase, considering that the shared features contain domain-relevant structural information, which is independent of category semantics, we design the Cross-Adaptive Modulation (CAM) module before the MGDF module, where shared features as prior knowledge are integrated into the private features through feature modulation techniques—enabling effective fusion of domain-relevant information and strengthening the domain adaptability. In summary, our contributions are:

\begin{itemize}
    \item We propose a novel CD-FSS method named DCDNet, which explicitly decouples base features and dynamically leverages these disentangled features to guide the construction of robust domain-invariant representations, effectively enhancing cross-domain generalization.
    \item We propose an innovative ACFD module with MGDF module to decouple base features, excavate latent information, and steer them toward domain-invariant representations; additionally, our CAM module modulates the decoupled features to ensure model effectiveness during fine-tuning adaptation.
    \item Extensive experiments conducted on four widely adopted target domain benchmarks demonstrate that our model achieves state-of-the-art performance among the leading methods in CD-FSS.
\end{itemize}

\section{Related Work}
\textbf{Few-Shot Semantic Segmentation.}
In recent years, with advances in computer vision and increasing requirements for recognizing novel categories with minimal data, the Few-Shot Segmentation (FSS)~\cite{peng2023hierarchical, fan2022self, lu2021simpler} has emerged, which has been studied extensively. The objective of FSS is to develop trained networks that can accurately recognize new categories with limited examples~\cite{allen2019infinite}. Most current approaches adhere to a meta-learning framework~\cite{vinyals2016matching}, which involves constructing learning episodes from base datasets by selecting support and query sets to simulate testing conditions, thereby allowing models to learn how to adapt to new segmentation tasks efficiently.
These FSS methods can be summarized in prototype-based learning and correlation-based learning. Prototype-based methods~\cite{li2024localization, liu2022dynamic, okazawa2022interclass} distill support information into single prototype or multiple prototypes, subsequently comparing or aggregating query features against these prototypes. Correlation-based methods~\cite{peng2023hierarchical, wang2023rethinking, xiong2022doubly} instead construct pixel-level correlations between support and query features to enable finer-grained interaction and integration. However, these conventional FSS methods are primarily designed for segmenting novel categories within the same domain. Their performance typically suffers significant degradation when applied across different domains.

\noindent \textbf{Cross-Domain Few-Shot Segmentation.} 
Unlike conventional FSS settings, Cross-Domain Few-Shot Segmentation (CD-FSS) requires models to generalize to unseen target domains without exposure to target data during training. Recently, CD-FSS has garnered increasing research attention. PATNet~\cite{lei2022cross} proposes a feature transformation layer that projects support and query features into a domain-agnostic space. DR-Adapter~\cite{su2024domain} employs a lightweight domain rectification adapter to align diverse target domain styles with the source domain. APM ~\cite{tong2024lightweight} proposes a Lightweight Frequency Masker to filter harmful frequency components at the feature level, thereby reducing inter-channel dependencies in feature maps and significantly enhancing performance and generalization capability. LoEC~\cite{liu2025devil} proposes a dual-stage collaborative mechanism addressing vulnerability in low-level features to enhance cross-domain generalization.
Most existing methods introduce adapter-like modules to backbone architectures or feature embeddings. However, we identify intrinsic information redundancy within base features themselves as a limitation. Instead of external adaptation, our approach decomposes and refines base features through structured disentanglement, then integrates these decoupled representations to endow models with enhanced generalization and domain adaptation capabilities.

\begin{figure*}[!t]
\centering
\includegraphics[width=\linewidth]{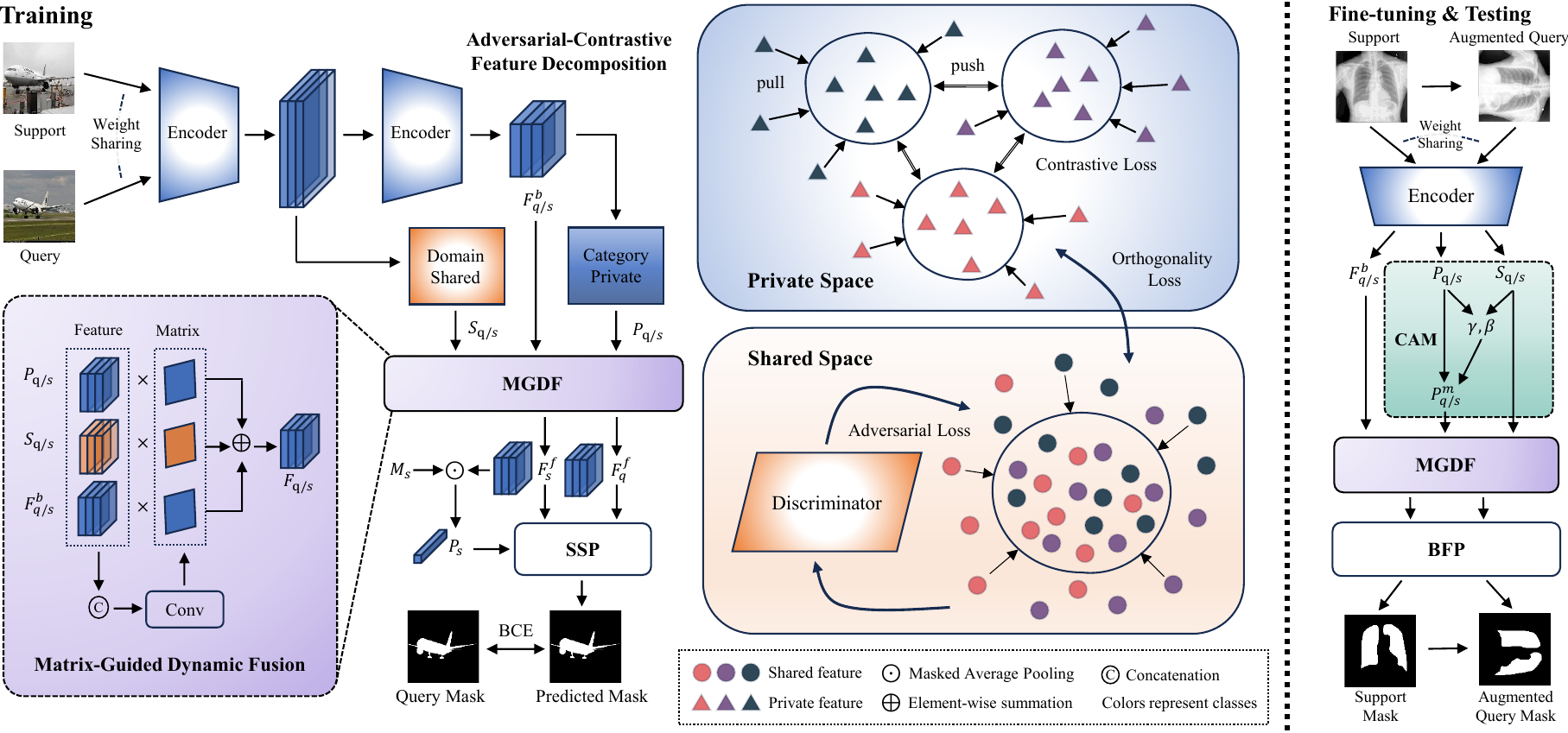}
\caption{\textbf{Overall architecture of our method in 1-shot example.} During training, feature maps are decomposed through the Adversarial-Contrastive Feature Decomposition module—leveraging adversarial and contrastive learning objectives—to yield disentangled representations of shared, private, and base features. These components subsequently undergo fusion via Matrix-Guided Dynamic Fusion, generating enhanced representations that drive query image segmentation through the SSP method~\cite{fan2022self}. During fine-tuning and testing, the Cross-Adaptive Modulation module strategically integrates prior knowledge from shared features into private features via feature modulation, while BFP module-based~\cite{nie2024cross} refinement further optimizes segmentation masks. }
\label{framework}
\end{figure*}

\section{Method}
\subsection{Problem Definition}
Cross-Domain Few-Shot Segmentation~\cite{lei2022cross} addresses the problem of transferring segmentation capabilities from a labeled source domain $\mathcal{D}_{\text{source}} = (\mathcal{X}^s, \mathcal{Y}^s)$ to an unseen target domain $\mathcal{D}_{\text{target}} = (\mathcal{X}^t, \mathcal{Y}^t)$ where $\mathcal{X}^s \neq \mathcal{X}^t$ (different data distributions) and $\mathcal{Y}^s \cap \mathcal{Y}^t = \emptyset$ (disjoint label spaces), using only $K$ annotated support examples $\mathcal{S} = \{(I_s^k, M_s^k)\}_{k=1}^K$ per novel category. Models are first meta-trained on source domain episodes $\mathcal{E} = (\mathcal{S}, \mathcal{Q})$ with both support masks $M_s^k$ and query masks $M_q$ available for supervision, then adapted to the target domain using only the support set $\mathcal{S}$ for potential model refinement without query mask access, and finally evaluated by segmenting target domain query images $I_q$ using only the support set $\mathcal{S}$ for conditioning with frozen model parameters during inference.

\subsection{Overview}

To enhance few-shot segmentation robustness while mitigating cross-domain interference, our framework employs the Adversarial-Contrastive Feature Decomposition module to disentangle domain-relevant shared features and category-relevant private features from backbone network. 
We then integrate category-relevant private features, domain-relevant shared features, and base features through the Matrix-Guided Dynamic Fusion module. Guided by a spatially-aware matrix, this fusion dynamically combines category details from private features with domain details from shared features and base features while preserving structural integrity of features.
During fine-tuning, we deploy the domain-relevant shared features to modulate category-relevant private features via lightweight affine transformations in the Cross-Adaptive Modulation module. This adaptive modulation mitigates the vulnerability of private features to cross-domain variations by injecting transferable domain-invariant knowledge.

\subsection{Training Phase}
\subsubsection{Adversarial-Contrastive Feature Decomposition Module}

In the CD-FSS tasks, multi-level backbone features often entangle domain-related and category-related information, which restricts generalization and hampers effective transfer and rapid adaptation to new domains. To end this, we design the Adversarial-Contrastive Feature Decomposition (ACFD) Module which leverages dual-branch structure and operates on different hierarchy levels of the backbone. As shown in Figure \ref{framework}, the shallow and deep features (\emph{i.e.,}  $\mathbf{F}_{q/s}^l$ and $\mathbf{F}_{q/s}^h$) are respectively fed into the shared and private networks to obtain shared features $\mathbf{S}_{q/s}$ and private features $\mathbf{P}_{q/s}$. To be specific, in the shared network, feature $\mathbf{F}_{q/s}^l$ is passed through two consecutive convolutional blocks and subsequently refined by an adaptive layer to enhance its structure representation,
\begin{equation}
\mathbf{S}_{q/s} = SA(\mathbf{F}_{q/s}^l) \times Conv_{3\times3}(Conv_{3\times3}(\mathbf{F}_{q/s}^l))
\end{equation}
where $\mathbf{F}_{q/s}^l \in \mathbb{R}^{C_{shared}\times H\times W}$ denotes backbone layer1 outputs, $Conv_{3\times3}$ denotes the convolution block which consists of $3\times3$ convolutional layer, instance normalization and ReLU operations, and $SA$ is the spatial attention mechanism. Simultaneously, in the private network, $\mathbf{F}_{q/s}^h$ undergoes a similar transformation pipeline as defined in shared network, with the key difference that the spatial attention module in the private network is replaced by a channel attention mechanism to enhance class-specific representations,
\begin{equation}
\mathbf{P}_{q/s} = CA(\mathbf{F}_{q/s}^h) \times Conv_{3\times3}(Conv_{3\times3}(\mathbf{F}_{q/s}^h))
\end{equation}
where $\mathbf{F}_{q/s}^h \in \mathbb{R}^{C_{private}\times H\times W}$ represents high-level features, $CA$ is the channel attention mechanism. 
However, directly using convolution and attention operations for the decomposition task cannot guarantee that the network learns to separate features based on domain and category information during training. 
Inspired by~\cite{ebrahimi2020adversarial, wang2022remember}, we employ adversarial learning to supervise the shared features. Specifically, a discriminator that can identify image categories for supervised training is leveraged, and with the help of Gradient Reversal Layer (GRL) ~\cite{ganin2016domain}, shared features are encouraged to align more closely with domain-relevant properties rather than being solely optimized for source domain segmentation, thereby enhancing its domain generalization capability for cross-domain scenarios, 
\begin{equation}
\mathcal{L}_{adv} = \frac{1}{N}\sum_{i=1}^N \left[\log D(\mathbf{S}(x_i^s)) + \log(1-D(\mathbf{S}(x_i^t)))\right]
\end{equation}
where $D$ denotes the classifier with architecture $D(\cdot) = MLP(GAP(\mathbf{S}))$. 
In addition, considering that optimizing the private branch solely through classification loss lacks an explicit mechanism to enhance intra-class compactness and inter-class separability, we adopt a contrastive learning paradigm to reconstruct and refine the private feature space. By forming positive pairs from same-class pixels to pull them closer and negative pairs from different classes to push them apart, the private features are guided to form compact, semantically coherent clusters with clearly defined boundaries, 
\begin{equation}
\mathcal{L}_{cont} = -\frac{1}{|\Omega|}\sum_{i\in\Omega}\log\frac{\exp(z_i\cdot z_{i^+}/\tau)}{\sum_{j\in\mathcal{N}(i)}\exp(z_i\cdot z_j/\tau)}
\end{equation}
where $\Omega$ denotes valid pixels, $z_i = Proj(P_i)$ the projected features, and $\mathcal{N}(i)$ the negative samples within a memory bank. 
Moreover, to ensure disentanglement completeness, we enforce orthogonality between shared and private features through batch-wise covariance minimization,
\begin{equation}
\mathcal{L}_{ortho} = \frac{1}{B}\sum_{b=1}^B\frac{\|\mathbf{S}_b^\top \mathbf{P}_b\|_F^2}{\|\mathbf{S}_b\|_F\|\mathbf{P}_b\|_F}
\end{equation}
where $S_b$ denotes the shared feature for the $b^{th}$ sample, $P_b$ represents the corresponding private feature, and $B$ is the batch size. Through this approach, the shared features are enabled to capture domain-relevant structural patterns, while the private features retain semantic-specific discriminative power, with their interaction dynamically regulated by orthogonality constraints.

\subsubsection{Matrix-Guided Dynamic Fusion Module}

To mitigate the potential information loss in the disentanglement, we propose the matrix-guided dynamic fusion method, as shown in Figure \ref{framework}. First, we integrate the base feature $\mathbf{F}_{q/s}^b$ extracted from the high-level feature with shared feature $\mathbf{S}_{q/s}$ and private feature $\mathbf{P}_{q/s}$ along the channel dimension and employ a 1$\times$1 convolution to reduce the dimensionality of the feature channels. In this way, the model enables interaction between shared domain information and private category information while preserving certain aspects of the original feature representation,
\begin{equation}
    \mathbf{F}_{q/s}^c = Conv_{1\times1}(Cat(\mathbf{F}_{q/s}^b,\mathbf{S}_{q/s},\mathbf{P}_{q/s}))
\end{equation}
where $Cat$ denotes the concatenation operation along the channel dimension, $Conv_{1\times1}$ is the 1$\times$1 convolutional layer. After that, in order to assign adaptive importance to different features based on their relevance, enabling the model to focus on more informative representations, we apply the 3$\times$3 convolutional layer and Softmax operation to the fusion feature $\mathbf{F}_{q/s}^c$ to generate three sets of feature-wise weights, which are then multiplied with features $\mathbf{F}_{q/s}^b$, $\mathbf{S}_{q/s}$, and $\mathbf{P}_{q/s}$ respectively to perform weighted fusion,
\begin{equation}
    \{w_b, w_s, w_p\} = Sp(\varphi (Conv_{3\times3}(\mathbf{F}_{q/s}^c))) \in \mathbb{R}^{1\times H\times W}
\end{equation}
where $\varphi$ denotes Softmax activation, $Sp$ denotes the split operation along the channel dimension. The final fused feature is obtained by summing the weighted features, 
\begin{equation}
    \mathbf{F}_{q/s}^f = {w_p\mathbf{P}_{q/s} + w_s\mathbf{S}_{q/s} + w_b\mathbf{F}_{q/s}^b} + \mathcal{G}(\mathbf{F}_{q/s}^c)
\end{equation}
where $\mathcal{G}(\cdot)$ denotes the enhancement block with 1×1 convolution and residual connection.
Finally, few-shot segmentation is performed using these fused features through the Self-Support Prototype (SSP) method ~\cite{fan2022self}: We first extract category prototypes from support features, predict initial segmentation masks based on query-prototype similarity, and iteratively refine predictions through query self-segmentation.

\subsection{Fine-tuning Phase}

\subsubsection{Cross-Adaptive Modulation Module}

To strengthen the domain adaptability, inspired by unsupervised RGB-D salient object detection algorithms~\cite{li2020rgb, feng2016local} that use depth maps as prior information to enrich saliency cues, we propose a novel Cross-Adaptive Modulation (CAM) module for domain adaptation during fine-tuning, which dynamically generates affine transformation parameters $\gamma$ and $\beta$ by modeling the interaction between private semantic features $\mathbf{P}_{q/s}$ and shared domain features $\mathbf{S}_{q/s}$, where shared features serve as prior knowledge to guide the modulation process, enabling private category features to rapidly adapt to target domain distributions while preserving their core semantic information. As shown in Figure \ref{framework}, the CAM module operates through three sequential operations, first, we implement cross-feature interaction by concatenating the private and shared features (\emph{i.e.,} $\mathbf{P}_{q/s}$ and $\mathbf{S}_{q/s}$) along the channel dimension, followed by a dimensionality-reducing convolution and a non-linear activation function,
\begin{equation}
    \mathbf{F}_{q/s}^a = ReLU(Conv_{3\times3}(Cat(\mathbf{S}_{q/s},\mathbf{P}_{q/s})))
\end{equation}
where $Conv_{3\times3}$ uses 3×3 kernels with padding=1 to preserve spatial dimensions, and ReLU activation $\max(0,x)$ enables effective cross-feature interaction learning. 
Next, we generate the adaptation parameters from these interaction features through a lightweight affine transformation, 
\begin{equation}
    \{\bm{\gamma},\ \bm{\beta}\} = tanh(Conv_{1\times1}(\mathbf{F}_{q/s}^a))
\end{equation}
here, the point-wise convolution $Conv_{1\times1}$ expands channels back to the original dimension $C$, while the $tanh$ activation constrains parameter values to the range $[-1,1]$ for stable gradient propagation. The scaling parameters $\gamma$ adaptively amplify or attenuate feature responses, while shifting parameters $\beta$ perform feature distribution translation.
After that, we modulate the private semantic features using these domain-relevant parameters,
\begin{equation}
\mathbf{P}_{q/s}^m = \mathbf{P}_{q/s} \times (\mathbf{1} + \bm{\gamma}) + \bm{\beta}
\end{equation}
By this way, CAM enhances discriminative robustness by transferring domain-relevant knowledge from shared to private features.
Finally, base feature $\mathbf{F}_{q/s}^b$, shared feature $\mathbf{S}_{q/s}$ and modulated private feature $\mathbf{P}_{q/s}^m$ are fused in the matrix-guided dynamic fusion module to generate feature $\mathbf{F}_{q/s}^f$.
Segmentation is ultimately achieved through the BFP method ~\cite{nie2024cross}. Building upon SSP, BFP introduces a prototype-mask iterative refinement mechanism: it first generates an initial segmentation mask, then cyclically performs two optimization steps—extracting query image prototypes based on the current mask, and subsequently fusing support prototypes with query prototypes through feature propagation to produce refined masks.

\subsection{Loss Function}
Our training protocol employs alternating optimization between the main segmentation model $\theta_d$ and the domain discriminator $\theta_a$. Each training iteration consists of two consecutive phases. First, during $s_{\text{steps}}$ update cycles, the main model minimizes a composite objective function combining multiple loss components,
\begin{equation}
\mathcal{L} = \lambda_{\text{ce}}\mathcal{L}_{\text{ce}} + \lambda_{\text{adv}}\mathcal{L}_{\text{adv}} + \lambda_{\text{cont}}\mathcal{L}_{\text{cont}} + \lambda_{\text{ortho}}\mathcal{L}_{\text{ortho}}
\end{equation}
where $\mathcal{L}_{\text{ce}}$ represents the supervised segmentation loss on source domain data, $\mathcal{L}_{\text{adv}}$ constitutes the feature-level adversarial loss promoting domain-relevant representations, $\mathcal{L}_{\text{orhto}}$ explicitly enforces distributional separation between domains, and $\mathcal{L}_{\text{contr}}$ facilitates semantic feature alignment across domains. Subsequently, during $d_{\text{steps}}$ update cycles, the discriminator minimizes its domain classification objective,
\begin{equation}
\mathcal{L}_{\text{disc}} = \mathcal{L}_{\text{real}} + \mathcal{L}_{\text{fake}}
\end{equation}
This discriminator loss encompasses $\mathcal{L}_{\text{real}}$ for authentic source feature classification and $\mathcal{L}_{\text{fake}}$ for detecting target features synthesized by the main model.

\section{Experiments}

\begin{table*}[t]
\setlength{\tabcolsep}{3pt}
 \renewcommand\arraystretch{1}
    \centering
    \begin{tabular}{c|c|c|cc|cc|cc|cc|cc}
        \toprule[1pt]
        \multirow{2}*{Methods} & \multirow{2}*{Backbone} & \multirow{2}*{Publication} & \multicolumn{2}{c|}{Deepglobe} & \multicolumn{2}{c|}{ISIC} & \multicolumn{2}{c|}{Chest X-Ray} & \multicolumn{2}{c|}{FSS-1000} & \multicolumn{2}{c}{\textbf{Average}} \\
        & & & 1-shot & 5-shot & 1-shot & 5-shot & 1-shot & 5-shot & 1-shot & 5-shot & 1-shot & 5-shot \\
        \midrule
        \multicolumn{13}{c}{Few-Shot Segmentation Methods} \\
        \midrule
        HSNet \shortcite{min2021hypercorrelation} & ResNet-50 & ICCV & 29.7 & 35.1 & 31.2 & 35.1 & 51.9 & 54.4 & 77.5 & 81.0 & 47.6 & 51.4\\
        SSP \shortcite{fan2022self} & ResNet-50 & ECCV & 40.0 & 48.7 & 35.5 & 45.9 & 74.4 & 74.3 & 78.9 & 80.6 & 57.2 & 62.4 \\
        PerSAM \shortcite{zhang2023personalize} & Vit-base & ICLR & 36.1 & 40.7 & 23.3 & 25.3 & 30.0 & 30.1 & 60.9 & 66.5 & 37.6 & 40.6\\
        \midrule
        \multicolumn{13}{c}{Cross-Domain Few-Shot Segmentation Methods} \\
        \midrule
        PATNet \shortcite{lei2022cross} & ResNet-50 & CVPR & 37.9 & 43.0 & 41.2 & 53.6 & 66.6 & 70.2 & 78.6 & 81.2 & 56.1 & 62.0 \\
        ABCDFSS \shortcite{herzog2024adapt} & ResNet-50 & CVPR & 42.6 & 49.0 & 45.7 & 53.3 & 79.8 & 81.4 & 74.6 & 76.2 & 60.7 & 65.0 \\
        DR-Adapter \shortcite{su2024domain} & ResNet-50 & CVPR & 41.3 & 40.1 & 40.8 & 48.9 & \underline{82.4} & 82.3 & 79.1 & 80.4 & 60.7 & 65.0 \\
        APSeg \shortcite{he2024apseg} & Vit-base & CVPR & 36.0 & 40.0 & 45.4 & 54.0 & \textbf{84.1} & \textbf{84.5} & 79.7 & 81.9 & 61.3 & 65.1 \\
        APM \shortcite{tong2024lightweight} & ResNet-50 & NeurIPS & 40.9 & 44.9 & 41.7 & 51.2 & 78.3 & \underline{82.8} & 79.3 & 81.9 & 60.0 & 65.2 \\
        IFA \shortcite{nie2024cross} & ResNet-50 & CVPR & \underline{50.6} & \underline{58.8} & \underline{66.3} & \underline{69.8} & 74.0 & 74.6 & \underline{80.1} & \underline{82.4} & \underline{67.8} & \underline{71.4} \\
        LoEC \shortcite{liu2025devil} & ResNet-50 & CVPR & 44.1 & 49.7 & 38.2 & 47.0 & 81.0 & 82.7 & 78.5 & 80.6 & 61.5 & 65.0 \\
        Ours & ResNet-50 &- & \textbf{51.3} & \textbf{62.5} & \textbf{72.0} & \textbf{79.8} & 80.7 & 81.1 & \textbf{81.7} & \textbf{83.3} & \textbf{71.4} & \textbf{76.7} \\
        \bottomrule[1pt]
    \end{tabular}
    \caption{Comparison of model performance in 1-shot and 5-shot settings. The best and second-best methods are highlighted in \textbf{bold} and \underline{underlined}, respectively.}
    \label{tab:performance}
\end{table*}

\subsection{Datasets and Implementation Details}
Following ~\cite{lei2022cross}, we evaluate our method on the Cross-Domain Few-Shot Segmentation (CD-FSS) benchmark, which comprises five datasets: PASCAL VOC 2012~\cite{everingham2010pascal}, FSS-1000~\cite{li2020fss}, DeepGlobe~\cite{demir2018deepglobe}, ISIC 2018~\cite{codella2019skin, tschandl2018ham10000}, and Chest X-Ray~\cite{candemir2013lung, jaeger2013automatic}. These datasets cover a wide range of domains—including natural scenes, satellite imagery, and medical diagnostics—providing substantial domain diversity for rigorous cross-domain evaluation. The model is trained exclusively on PASCAL VOC 2012, augmented with the SBD dataset~\cite{hariharan2011semantic}, which serves as the source domain containing 20 common object classes. It is then fine-tuned and evaluated on the remaining four target-domain datasets. FSS-1000 comprises 1,000 natural image categories, primarily featuring rare or fine-grained object classes. DeepGlobe includes seven land-cover types—urban, agriculture, rangeland, forest, water, barren, and unknown. Following~\cite{lei2022cross}, we preprocess the satellite images by dividing them into smaller patches and excluding samples labeled as “unknown” prior to training. ISIC 2018 provides dermoscopic images for the classification of three types of skin lesions. Lastly, the Chest X-ray dataset consists of grayscale images specifically collected from patients with tuberculosis.

We begin by pretraining SSP~\cite{fan2022self} on the PASCAL VOC dataset, designated as the source domain, using the episodic training paradigm. After pretraining, the backbone is frozen, and our method is trained on it for 20 epochs with a batch size of 8. All training was performed on an NVIDIA RTX 3090 GPU. The primary model is optimized using stochastic gradient descent (SGD) with a momentum of 0.9 and an initial learning rate of 1e-3, while the domain discriminator is trained using the Adam optimizer with a weight decay of 0.01 and an initial learning rate of 1e-4. To reduce memory consumption and computational cost, both support and query images are resized to $400 \times 400$ pixels~\cite{lei2022cross}. During fine-tuning, the learning rate is set to 5e-4 for the DeepGlobe, ISIC, and FSS-1000 datasets, and to 1e-5 for the Chest X-ray dataset, with all target-domain models trained for 40 epochs. Support images are augmented using randomized transformations, including horizontal and vertical flipping, 90-degree rotations, brightness adjustments, and hue variations, following ~\cite{nie2024cross}. Model performance is evaluated using the mean Intersection over Union (mIoU)~\cite{min2021hypercorrelation}.

\begin{figure}[!t]
\centering
\includegraphics[width=0.46\textwidth]{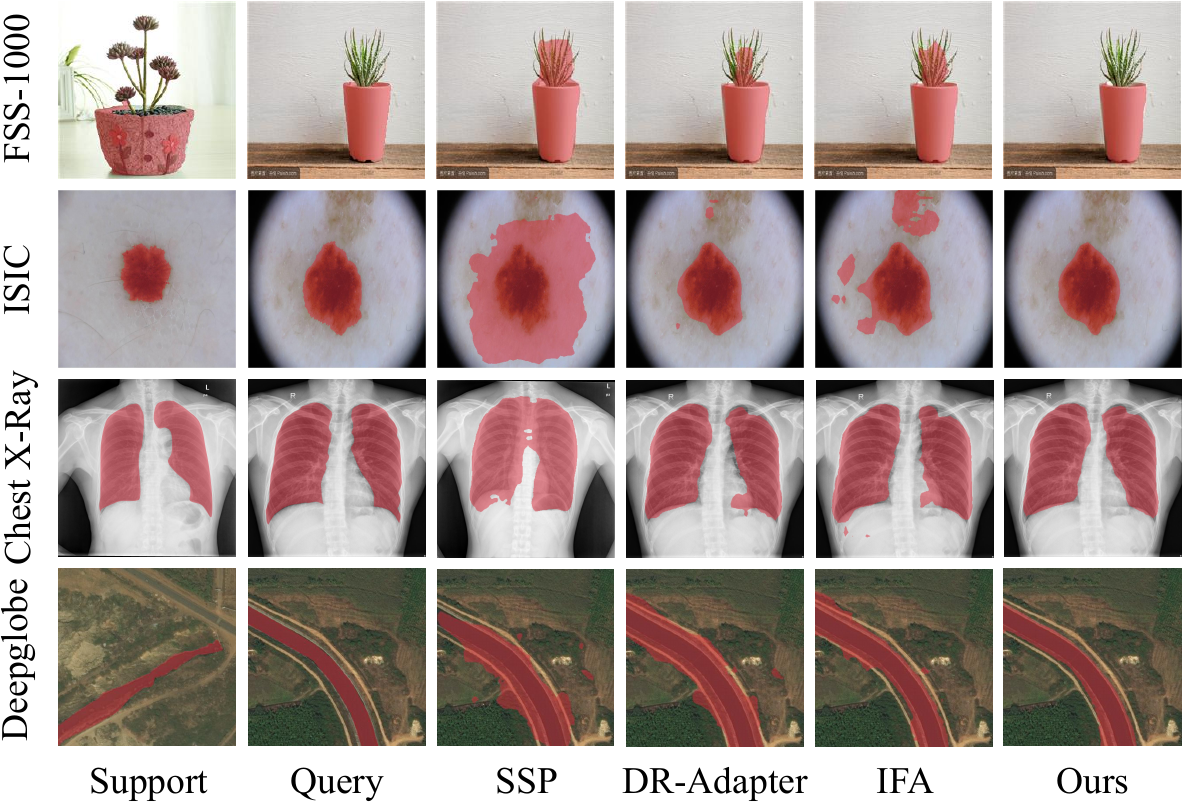}
\caption{Qualitative results of the samples in four target datasets. From up to down, each row shows examples from FSS-1000, ISIC, Chest X-Ray, and Deepglobe. From left to right, each column shows the examples of support images with ground-truth masks, query images with ground-truth masks, SSP results, DR-Adapter results, IFA results and Our results.}
\label{vis}
\end{figure}

\subsection{Comparison with State-of-the-art Methods}

As shown in Table \ref{tab:performance}, we conducted a comprehensive comparison between our Divide-and-Conquer Decoupled Network (DCDNet) and state-of-the-art methods using mean Intersection over Union (mIoU) as the evaluation metric. To ensure a fair comparison, we adopt the same feature extraction network, ResNet-50~\cite{he2016deep}, as used in most baseline methods. The numerical results in Table \ref{tab:performance} demonstrate that our method achieves substantial performance gains across diverse datasets, where DCDNet achieves superior detection performance on the ISIC dataset, with mIoU scores of $72.0\%$ (1-shot) and $79.8\%$ (5-shot), surpassing the previous best method ~\cite{nie2024cross} by $+8.6\%$ and $+14.3\%$, respectively. 
From the above comparison results, we can see that our model demonstrates outstanding cross-domain adaptability, effectively handling medical images with complex pathological structures despite the significant domain gap from natural images. 
In addition, for the challenging DeepGlobe satellite dataset - characterized by category diversity and densely annotated scenes where extraneous classes often interfere with background segmentation - DCDNet also achieves remarkable performance of 51.3\% (1-shot) and 62.5\% mIoU (5-shot), demonstrating that the robustness of our method in complex, multi-category environments where other traditional approaches are prone to category confusion. Besides, on medical imaging datasets (\emph{i.e.,} ISIC and Chest X-Ray), which typically feature uniform backgrounds but contain subtle pathological details with significant domain gaps from natural images, our model maintains competitive performance. The above comparison results suggest that the proposed disentanglement strategy effectively preserves essential diagnostic features under domain shift, as evidenced by its superior performance across diverse medical imaging benchmarks. Moreover, on the FSS-1000 dataset which contains rare categories and small objects, our method achieves state-of-the-art performance of 81.7\% mIoU (1-shot) and 83.3\% mIoU (5-shot), demonstrating its effectiveness not only in addressing cross-domain challenges but also in maintaining strong performance on complex in-domain few-shot segmentation tasks. Finally, to more comprehensively illustrate the overall effectiveness of the proposed method, we calculate the average performance across all four datasets, as reported in the last column of Table \ref{tab:performance}, where our method achieves average mIoU improvements of +5.3\% (1-shot) and +7.4\% (5-shot) over the previous state-of-the-art, establishing new benchmarks for CD-FSS. To better analyze and understand the superiority of DCDNet, we visualize the segmentation results, as shown in Figure \ref{vis}. Our results ($6^{th}$ column) are significantly better compared to the previous best methods ($4^{th}$ and $5^{th}$ column). We also visualize more qualitative results of DCDNet in supplementary materials.

\subsection{Ablation Studies}

\begin{table}[!t] \footnotesize
  \centering  
  \begin{threeparttable}
    \begin{tabular}{>{\centering\arraybackslash}p{1.2cm}
                    >{\centering\arraybackslash}p{1.2cm}
                    >{\centering\arraybackslash}p{1.2cm}|
                    >{\centering\arraybackslash}p{1.5cm}}
    \toprule  
    MGDF&ACFD&CAM&mIoU (\%)\cr
    \midrule
         &     &     &  80.1\cr
    \checkmark     &     &     &   80.6\cr
    \checkmark     &\checkmark     &     &  81.3\cr
    \checkmark     &\checkmark     &\checkmark     &  \textbf{81.7}\cr
    \bottomrule  
    \end{tabular}  
    \end{threeparttable}  
    \caption{Effectiveness of MGDF,  ACFD and CAM modules.} 
    \label{tab:module}
\end{table}

\noindent \textbf{Effects of different modules.} As shown in Table \ref{tab:module}, we conducted four ablation experiments on FSS datasets by incrementally adding three modules to verify the effectiveness of each key component (\emph{i.e.,} MGDF, ACFD and CAM) in our DCDNet. To be specific, for the 1$^{st}$ row, we conducted a baseline where all three modules were removed, and leveraged features extracted from the backbone network to execute prototype-based few-shot segmentation via the SSP method~\cite{fan2022self}, followed by fine-tuning aligned with the IFA optimization framework ~\cite{nie2024cross}. After that, to verify the effectiveness of MGDF module, we added the MGDF module on the basis of the baseline to fuse the private and shared features, as shown in the 2$^{nd}$ row. From the comparison results in the Table \ref{tab:module}, it can be seen that compared to the baseline, the addition of MGDF module has improved the mIoU to $ 80.6\%$. Next, we added the ACFD module on the basis of the above experiment ( 3$^{rd}$ row), where backbone features are decomposed into shared and private features which are supervised by adversarial and contrastive learning before the MGDF module. From the comparison results in Table \ref{tab:module}, it can be seen that the utilize of ACFD module reduces cross-domain interference and enhances model generalization and domain adaptation capabilities, and significantly improved the detection metric. Finally, to evaluate the effectiveness of the CAM module, we further integrated it into the network, where the private features were modulated by the shared features during fine-tuning. The comparison between the fourth and third experiments demonstrates that the CAM module significantly enhances performance in the CD-FSS task.


\noindent \textbf{Effects of feature decomposition.} To evaluate the impact of feature decomposition on the model's detection performance, we conducted three separate experiments, as shown in Table \ref{tab:feature}. First, for the experiment 1 (1$^{st}$ row), we conducted a baseline which just leverages the base feature to generate the detection results. In addition, for the experiment 2 (2$^{nd}$ row), we decoupled the backbone features into category-relevant private and domain-relevant shared features, which are fed into the MGDF module for completing the feature fusion. As evidenced by the results in the Table \ref{tab:feature}, the model’s performance is significantly improved after applying feature disentanglement. Moreover, for the experiment 3 (3$^{rd}$ row),  we further introduce the base feature to ensure that a certain amount of original feature information is retained during the fusion of private and shared features. 
Overall, these experiments suggest that while the private feature segments well, its cross-domain generalization is limited; the shared feature, despite its weak standalone segmentation ability, offers essential complementary information; and the base feature plays a critical role in compensating for information loss during multi-feature integration. 
Collectively, our experimental results validate the effectiveness of the decomposed feature fusion process and the essential role of the base feature in supplementing information during integration.

\begin{table}[!t] \footnotesize
  \centering  
  \begin{threeparttable}
    \begin{tabular}{>{\centering\arraybackslash}p{1.2cm}
                    >{\centering\arraybackslash}p{1.2cm}
                    >{\centering\arraybackslash}p{1.2cm}|
                    >{\centering\arraybackslash}p{1.5cm}}
    \toprule  
    base&private&shared&mIoU (\%)\cr
    \midrule
    \checkmark     &     &      &  80.6\cr
         &\checkmark     &\checkmark     &  81.4\cr
    \checkmark     &\checkmark     &\checkmark     &  \textbf{81.7}\cr
    \bottomrule  
    \end{tabular}  
    \end{threeparttable}  
    \caption{Effectiveness of feature decomposition.} 
    \label{tab:feature}
\end{table}

\noindent \textbf{Effects of different loss.}
As shown in Table \ref{tab:loss}, we systematically decouple three loss components to investigate the influence of different supervisory mechanisms on model performance. The experimental setup begins with a baseline that includes only the segmentation loss (\emph{i.e.,} bce loss). Subsequently, introducing either the adversarial loss or the contrastive loss individually yields only modest improvements, while applying both together leads to a significant performance boost. Finally, we conduct an experiment which integrates three loss functions (\emph{i.e.,} adversarial loss, contrastive loss, and orthogonality loss) into the network, as shown in Table \ref{tab:loss} (5$^{th}$ row). From the above results, it is evident that the synergy among multiple supervisory signals, especially the combined effect of adversarial and contrastive losses, plays a critical role in unlocking the model’s potential, with the complete loss design ensuring optimal performance through complementary supervision.

\begin{table}[!t] \footnotesize
  \centering  
  \begin{threeparttable}
    \begin{tabular}{>{\centering\arraybackslash}p{1.2cm}
                    >{\centering\arraybackslash}p{1.2cm}
                    >{\centering\arraybackslash}p{1.2cm}|
                    >{\centering\arraybackslash}p{1.5cm}}
    \toprule  
    adv&cont&ortho&mIoU (\%)\cr
    \midrule
         &      &      &  80.6\cr
    \checkmark     &     &      &   81.0\cr
         &\checkmark     &      &  81.1\cr
    \checkmark     &\checkmark     &      &  81.5\cr
    \checkmark     &\checkmark     &   \checkmark   &  \textbf{81.7}\cr
    \bottomrule  
    \end{tabular}  
    \end{threeparttable}  
    \caption{Effectiveness of adversarial loss, contrastive loss and orthogonal loss.} 
    \label{tab:loss}
\end{table}

\section{Conclusion}
In this paper, we propose the Divide-and-Conquer Decoupled Network (DCDNet) to address CD-FSS task.
Specifically, in the training phase, our method decomposes encoder features into two features via adversarial and contrastive learning: category-relevant private features preserving high-level semantics and domain-relevant shared features capturing structural patterns. 
To address the potential degradation introduced by the disentanglement, we additionally fuse base features from the backbone with the modulated features through a spatial-level, matrix-guided strategy. Finally, during fine-tuning adaptation, we leverage shared features to modulate private features in the target domain, enabling efficient and lightweight adaptation with limited target-domain samples. Extensive experiments demonstrates that our DCDNet achieves state-of-the-art performance on various CD-FSS datasets.

\section{Acknowledgments}
This work was supported in part by the opening project of State Key Laboratory of Autonomous Intelligent Unmanned Systems under Grant ZZKF2025-2-8, in part by the National Natural Science Foundation of China under Grant 62471278 and 62271180, in part by the Key R\&D Program of Shandong Province under Grant 2025CXGC010111, and in part by the Taishan Scholar Project of Shandong Province under Grant tsqn202306079.

\bibliography{ref}

\end{document}